\documentclass{IEEEtran}
\usepackage{graphicx}
\usepackage{amsmath}
\usepackage{cite}
\usepackage{booktabs}

\usepackage{cite} 

\usepackage{array}
\usepackage{float}
\usepackage{caption}
\usepackage[table]{xcolor}
\usepackage{xcolor}
\usepackage[linkcolor=blue]{hyperref}
\usepackage{wrapfig}
\usepackage[symbol]{footmisc}

\title{Cached Adaptive Token Merging: Dynamic Token Reduction and Redundant Computation Elimination in Diffusion Model}
\author{Omid Saghatchian$^1$\textsuperscript{*} , Atiyeh Gh. Moghadam$^2$\textsuperscript{*}, Ahmad Nickabadi$^2$\\
    $^1$Mathematics and Computer Science, Amirkabir University of Technology, Tehran, Iran\\
    $^2$Computer Engineering, Amirkabir University of Technology, Tehran, Iran\\
    \{omidsaghatchian, atiyeh.moghadam, nickabadi\}@aut.ac.ir}
\begin{document}
\maketitle

\begin{abstract}
Diffusion models\footnotetext{$^*$These authors contributed equally to this work.}  have emerged as a promising approach for generating high-quality, high-dimensional images. Nevertheless, these models are hindered by their high computational cost and slow inference, partly due to the quadratic computational complexity of the self-attention mechanisms with respect to input size. Various approaches have been proposed to address this drawback. One such approach focuses on reducing the number of tokens fed into the self-attention, known as token merging (ToMe). In our method, called Cached Adaptive Token Merging (CA-ToMe), we calculate the similarity between tokens and then merge those tokens that have a similarity greater than a threshold parameter 
$t$. However, due to the repetitive patterns observed in adjacent steps and the variation in the frequency of similarities, we aim to enhance this approach by implementing an adaptive threshold for merging tokens and adding a caching mechanism that stores similar pairs across several adjacent steps. Empirical results demonstrate that our method operates as a training-free acceleration method, achieving a speedup factor of 1.24 in the denoising process while maintaining the same FID scores compared to existing approaches. Our code is available at \textcolor[rgb]{0.2,0.5,0.8}{\url{https://github.com/omidiu/ca_tome}}

\end{abstract}

%

\section{Introduction}
Over the past few years, diffusion models [\citenum{dhariwal2021diffusion}][\citenum{ho2020denoising}][\citenum{sohl2015deep}] have undergone a significant transformation, emerging as a notable advancement in the domain of generative modeling and garnering considerable attention from researchers in the field. These models have demonstrated substantial efficacy across various applications, including the generation of images[\citenum{kawar2022denoising}][\citenum{rombach2022high}], text[\citenum{gong2022diffuseq}][\citenum{li2022diffusion}], audio[\citenum{kong2020diffwave}][\citenum{popov2021grad}], and video[\citenum{luo2023videofusion}][\citenum{ho2022imagen}], as well as downstream tasks such as super-resolution[\citenum{li2022srdiff}][\citenum{gao2023implicit}], editing[\citenum{brooks2023instructpix2pix}][\citenum{avrahami2022blended}], and text-to-3D image generation[\citenum{poole2022dreamfusion}][\citenum{singer2022make}]. Notably, several of these models have been pre-trained on vast datasets comprising billions of text-image pairs, referred to as foundation models. In the realm of text-to-image generation, prominent examples of these foundation models include LDM[\citenum{rombach2022high}], DALL-E[\citenum{ramesh2022hierarchical}], Imagen[\citenum{saharia2022photorealistic}], and EMU[\citenum{dai2023emu}], which possess the remarkable capability of generating high-quality, photorealistic images based on user input prompts.

\begin{figure}[htbp]
    \centering
    \includegraphics[width=0.5\textwidth]{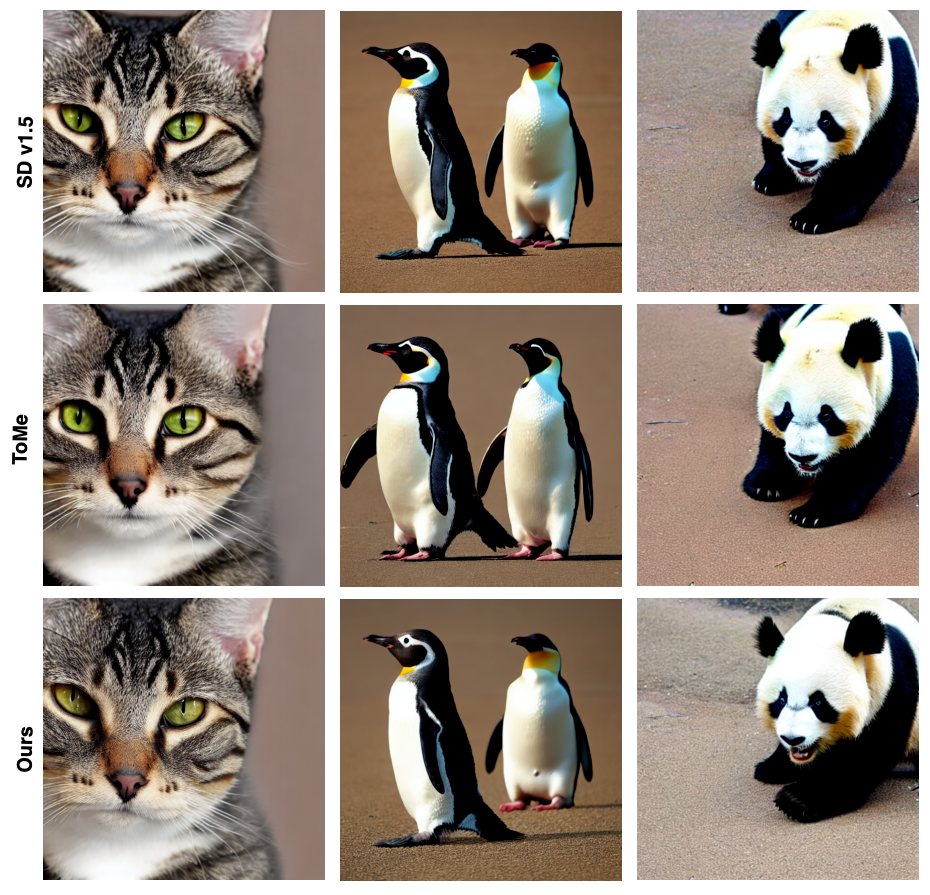}
    \caption{A comparison between the images generated by three different models. (First Row) Examples of images generated using SDv1.5[\citenum{rombach2022high}] without any token reduction, (Second row) Token merging[\citenum{bolya2022token}] with $r = 50\%$. (Third row) Our method(CA-ToMe) with $Threshold=0.7$.This comparison illustrates that our method works better in background details.}
    \label{fig:example}
\end{figure}

Despite the impressive performance of diffusion models, a major limitation of these models is their high latency and computational cost, which results in slow inference time. This latency issue renders diffusion models unsuitable for applications that require rapid and frequent inference. As noted in [\citenum{li2024snapfusion}], the primary challenge inherent to diffusion models lies in the slow and sequential denoising process, which is inherently serial and cannot be parallelized. Furthermore, each step of the denoising network, typically consisting of a U-Net[\citenum{ronneberger2015u}] architecture with residual connections and transformer blocks, is computationally intensive and thus cannot be evaluated repeatedly without incurring significant delays.

According to [\citenum{ma2024deepcache}], two primary strategies have been proposed to accelerate inference in diffusion models: reducing the number of sampling steps[\citenum{salimans2016improved}][\citenum{salimans2022progressive}][\citenum{lu2022dpm}][\citenum{meng2023distillation}] and diminishing the computational requirements of each denoising step[\citenum{zhang2024laptop}][\citenum{kim2023bk}][\citenum{he2024ptqd}]. The first approach focuses on minimizing the number of denoising steps required. A significant body of research has concentrated on developing novel samplers using numerical methods for solving ordinary differential equations (ODEs) and stochastic differential equations (SDEs), as they often conceptualize the generation process as an SDE or ODE[\citenum{song2020denoising}][\citenum{liu2022pseudo}][\citenum{chung2022come}]. Knowledge distillation is another dominant theme in this approach, aiming to reduce the number of steps using a student network that learns from the original denoising network[\citenum{salimans2022progressive}][\citenum{luhman2021knowledge}]. However, while new samplers often prevent the need for retraining, most other methods within this approach necessitate retraining, thereby having high computational costs. The second approach seeks to reduce the computational load at each step by examining and optimizing the various components of the denoising network. Techniques such as quantization[\citenum{he2024ptqd}][\citenum{shang2023post}], caching[\citenum{ma2024deepcache}][\citenum{wimbauer2024cache}][\citenum{zhang2024cross}][\citenum{li2023faster}], and token reduction[\citenum{bolya2022token}][\citenum{bolya2023token}][\citenum{yin2022vit}] are notable methods within this framework. A key advantage of these methods lies in their simplicity of implementation and the absence of retraining requirements.

The main goal of this article is to speed up the inference process by reducing the amount of computation required at each step. We achieve this by reducing the number of tokens used to evaluate transformers. Typically, each transformer block in the denoising U-net considers the pixels of the image in the latent space to be a token and processes every input token in each denoising step. However, most images, including those generated during the denoising process, contain redundant information. We can decrease the computational cost by reducing the number of tokens and skipping unnecessary computations. According to [\citenum{kim2024token}], two primary approaches have been identified to reduce the number of tokens: token pruning[\citenum{yin2022vit}][\citenum{rao2021dynamicvit}] and token merging[\citenum{bolya2022token}][\citenum{bolya2023token}][\citenum{marin2021token}]. Token pruning entails omitting a subset of tokens that have a negligible impact on the overall performance of the transformer. However, the main drawbacks of this method are the need to train the model to select tokens for pruning and the inevitable loss of information and performance degradation resulting from pruning many tokens. In contrast, token merging involves combining similar tokens instead of pruning, which removes them. Notably, token merging has been successfully implemented on diffusion models without the need for retraining and yielding significant reductions in inference time[\citenum{bolya2023token}].


Notwithstanding the benefits of the token merging method, it possesses several drawbacks. The primary limitation of this approach lies in its tendency to use a constant merging rate in order to merge tokens at each block and timestep. While this approach may have been viable if we were to consider U-Net as a black box, previous studies [\citenum{ma2024deepcache}][\citenum{wimbauer2024cache}] have highlighted that changes in different blocks at various timesteps exhibit smooth and identical patterns. Leveraging this insight, our method seeks to introduce adaptivity into the merging process by employing different merging rates. Furthermore, during the development of our method, we encountered a computational bottleneck: each time tokens are merged, a similarity matrix must be recalculated, resulting in an overhead of quadratic calculations. To mitigate this issue, we draw inspiration from [\citenum{ma2024deepcache}][\citenum{wimbauer2024cache}][\citenum{li2023faster}] and explore the possibility of caching calculations to reduce the computational cost of similarity matrix computation. Our experimental results demonstrate that our approach can reduce image generation time by a factor of 1.24 while only marginally increasing the FID, which is less than the increase observed in the baseline token merging model.

In the subsequent sections, we will first conduct a comprehensive review of existing methods for accelerating diffusion models, focusing on token reduction techniques in the background section. This will be followed by Section III, in which we will outline our novel contributions and elucidate the underlying motivations and inspirations that led to these ideas. The implementation details and experimental setup will be presented in Sections IV and V, respectively.

\begin{figure*}[htbp]
    \centering
    \includegraphics[width=1\textwidth]{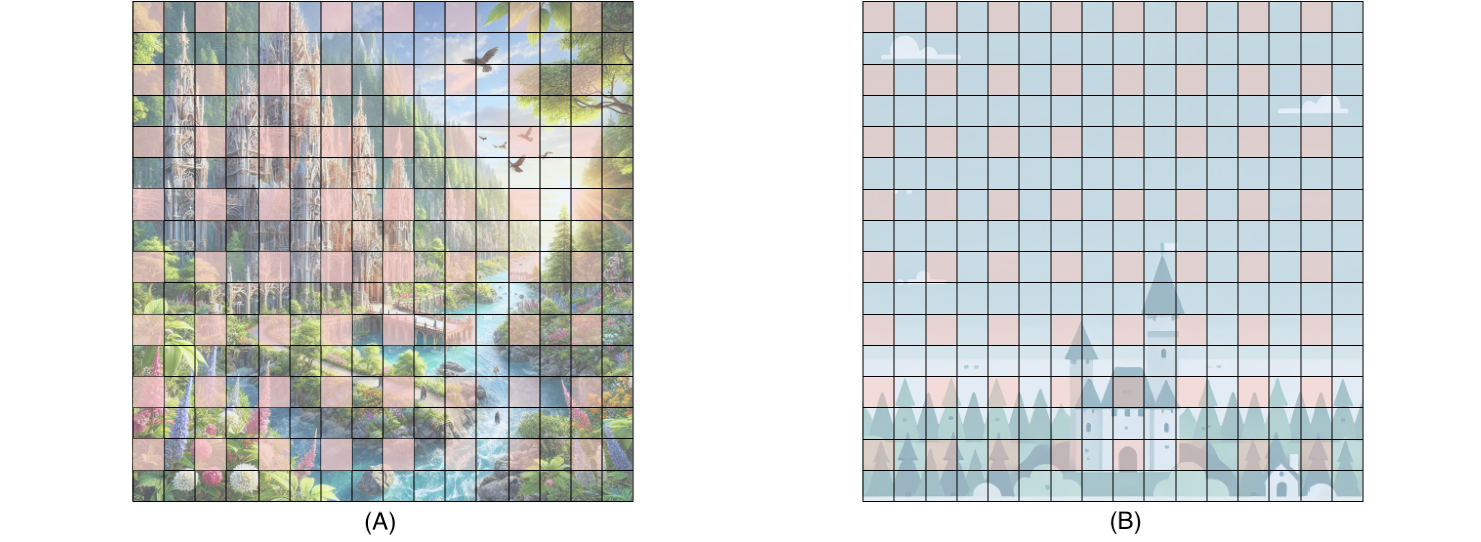}
    \caption{This figure shows a comparison between different scenarios in the frequency of similarity values in a self-attention block. The plots show the histogram of similarities between input tokens. The blue bars show the similarities that aren’t in the $r$ most similar tokens so that they won’t be merged. \textbf{(A)} Shows the scenario where most of the tokens are not similar, so when we use a constant merging rate, we are allowing our method to merge dissimilar tokens, which can lead to information loss. However, when we determine a threshold, as shown in the figure, it tends to merge fewer tokens and only select similar ones. \textbf{(B)} Shows the scenario where most of the tokens are similar(A typical scenario in the first steps of denoising). If we use a constant merging rate, we are forcing our method to choose tokens that are less similar than it could choose without any damage to quality. Again, in this scenario, selecting a threshold for merging can lead to merging more tokens and speeding up the inference.}
    \label{fig:adaptive}
\end{figure*}

\section{Background}
The generation of high-dimensional high-quality images has been an active area of research with early models such as Generative Adversarial Networks(GANs)[\citenum{goodfellow2014generative}][\citenum{arjovsky2017wasserstein}][\citenum{li2019controllable}] and Variational Autoencoders(VAEs)[\citenum{higgins2017beta}][\citenum{kingma2013auto}] facing limitations due to instability and mode collapse which hindered their scalability. In contrast, diffusion models have demonstrated the ability to generate high-quality images without these challenges. However, the denoising process intrinsic to diffusion models is computationally expensive resulting in slow inference.

Diffusion models operate by incrementally refining pure noise to produce a realistic image. Over time, different implementations of diffusion models have emerged, utilizing various architectures and components. Examples of these implementations include EMU[\citenum{dai2023emu}], DIT[\citenum{peebles2023scalable}], and Stable Diffusion[\citenum{rombach2022high}]. Stable diffusion models adopt a similar approach, albeit in the latent space. The stable diffusion network architecture consists of three primary components: a text encoder, which converts input text into an embedding; a Variational Autoencoder(VAE), which maps between the image and latent spaces; and a denoising network, comprising a U-Net with residual blocks and transformer blocks. In this framework, each image in the latent space is decomposed into a set of tokens, which are then fed into a series of downsampling and upsampling blocks in U-Net to compute the output of the denoising network.

\begin{figure*}[htbp]
    \centering
    \includegraphics[width=1\textwidth]{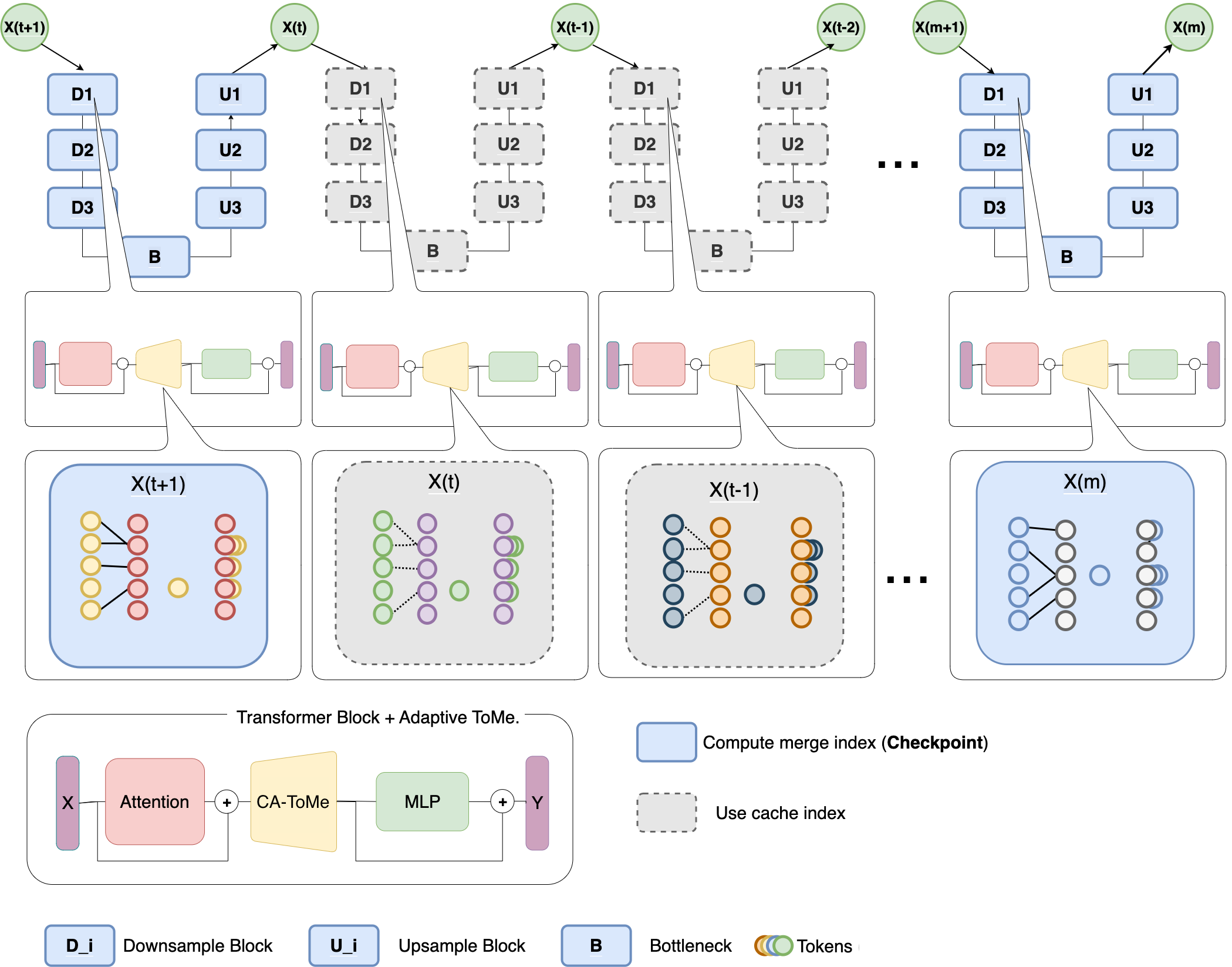}
    \caption{This figure demonstrates the whole scheme of pair caching. The above U-shaped blocks show denoising U-nets with the upblocks and downblocks. In each of these blocks, there exist some transformer blocks containing attention mechanisms. In some timesteps, which are illustrated with blue boxes, we calculate the whole process of token merging, but in other timesteps, which are illustrated with gray boxes, we just use the same pairs from the previous timestep to do token merging. In both the blue and gray boxes, there exists a bipartite graph where the left side represents source tokens and the right side represents destination tokens, depicted in different colors. Additionally, the different colors of tokens across the boxes indicate that they do not have the same values.}
    \label{fig:graphical_abs}
\end{figure*}

Two primary strategies have been identified to enhance inference speed and reduce latency. The first approach involves diminishing the number of inference iterations, which can be accomplished through various methods, including knowledge distillation [\citenum{salimans2022progressive}][\citenum{huang2024knowledge}][\citenum{meng2023distillation}], implicit sampling [\citenum{song2020denoising}], advanced differential equation solvers [\citenum{bao2022analytic}][\citenum{liu2022pseudo}], and parallel sampling [\citenum{shih2024parallel}]. Most of these methods require extensive changes to different parts of the network and sampling process, necessitating retraining. However, since the cost of training a diffusion model on datasets like ImageNet[\citenum{deng2009imagenet}] can be prohibitively high, it is not feasible for many researchers to utilize these approaches.

Conversely, numerous studies have focused on the second strategy, which aims to reduce computations within each step. This approach is primarily realized through network compression, which can be achieved via several techniques, including quantization [\citenum{he2024ptqd}][\citenum{shang2023post}] and token reduction [\citenum{bolya2022token}][\citenum{bolya2023token}][\citenum{yin2022vit}][\citenum{rao2021dynamicvit}][\citenum{marin2021token}]. Additionally, a recent paradigm has emerged, involving the caching of distinct components of the denoising network [\citenum{ma2024deepcache}][\citenum{wimbauer2024cache}][\citenum{zhang2024cross}][\citenum{li2023faster}]. These approaches are easy to implement and work in a plug-and-play manner. Furthermore, they possess the added advantages of simplicity and flexibility, allowing them to be applied to a variety of architectures and implementations of diffusion models. Notably, they have been shown to reduce inference time by half in some approaches, with little to no degradation in image quality.

The token merging method [\citenum{bolya2023token}] is a strategy aimed at reducing the computational complexity inherent in each timestep. Notably, since transformers exhibit quadratic scaling in terms of input tokens, decreasing the number of input tokens can lead to a significant reduction in computations per step of the denoising process. The token merging method operates by partitioning the input tokens of the transformer into source and destination sets, thereby creating a bipartite graph. It then calculates the similarity using the cosine similarity metric to find the top $r$ proportion of most similar tokens in source and destination sets for merging purposes. However, we observe that there are patterns in similarity between tokens that change with respect to different timesteps. Considering a rigid constant for merging can degrade the method's capabilities. Therefore, we propose making the merging rate $r$ adaptive to better capture these varying patterns.

Recently, caching has emerged as a promising approach to reduce the computational cost of diffusion models. This technique leverages previously computed results to avoid redundant calculations at each timestep. Existing caching algorithms target various components of denoising networks, including cross-attention [\citenum{zhang2024cross}], U-Net encoders [\citenum{li2023faster}], and high-level features [\citenum{ma2024deepcache}]. While these methods effectively reduce computational cost and time, some compromise image quality. These methods mainly use the smoothness of changes in timesteps and the similar patterns that can be found in different blocks in adjacent timesteps to skip some calculations. Inspired by this and recognizing the computational burden of repeatedly calculating the similarity matrix within our merging function, we propose caching this matrix, computing it only at selected timesteps rather than at every step.

\section{Methodology}
Our objective is to produce a high-quality image in less time. We addressed quality and speed separately, applying distinct approaches for each and using them together for optimal performance. The first approach is adaptive token merging, which performs the merging operation based on the similarity distribution among tokens and improves the overall generated image quality. The second approach involves caching part of the computations, which boosts processing speed. In Sections \ref{subsec:adaptive} and \ref{subsec:caching}, we will explore each approach in detail. 

Our approach explicitly targets the transformer blocks within the uppermost layers of the U-Net architecture, including both the highest encoder and decoder layers, $D_1$ and $U_1$, because the higher number of tokens in these layers makes merging more beneficial. Merging would not significantly improve performance in the lower layers, with fewer tokens. According to [\citenum{li2024snapfusion}]. In transformer blocks, self-attention is the most computationally expensive block, so we seek to focus on this block and decline the number of tokens in it. Modifying the computations within these blocks and caching specific values across different inference steps significantly increases the model’s processing speed without compromising output quality\ref{tab:final_results}.

\begin{figure*}[htbp]
    \centering
    \includegraphics[width=1\textwidth]{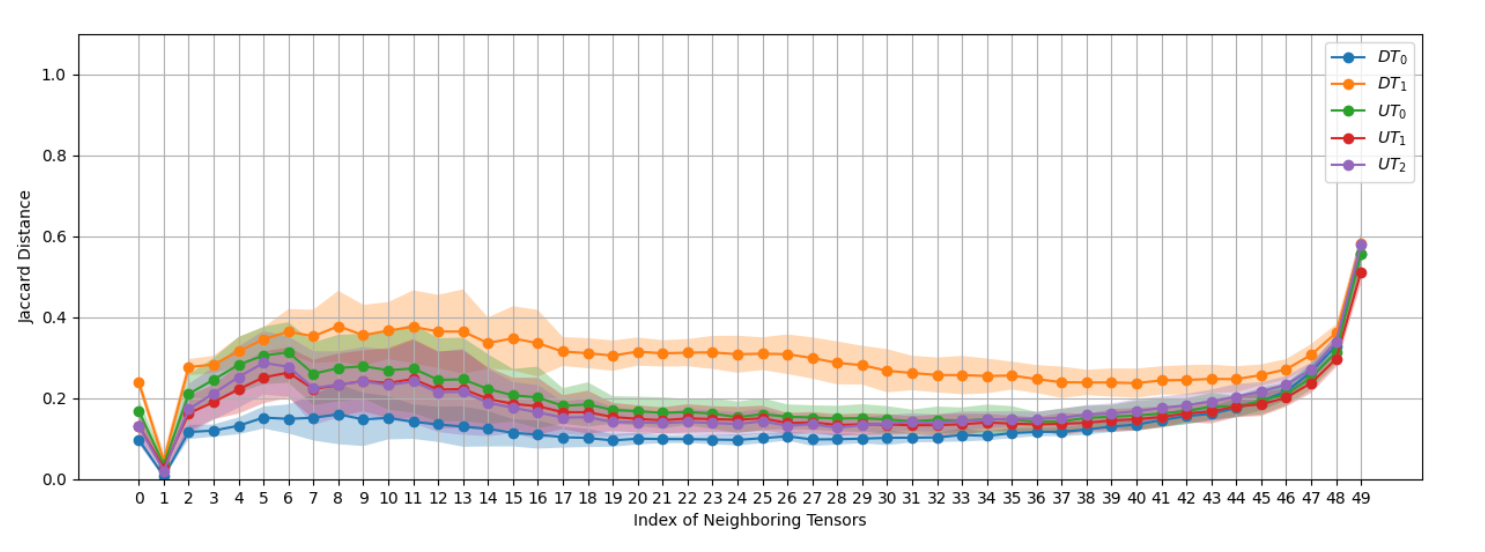}
    \caption{The Jaccard distance between pairs in adjacent steps across different transformer layers within $D_1$ and $U_1$ blocks. This figure is plotted using 100 photos generated using 100 classes of ImageNet. The shaded regions around the curves represent the variance in the Jaccard distance across these images. As illustrated in the figure, most intermediate steps exhibit a distance of less than 0.2, indicating high similarity among pair sets.}
    \label{fig:jaccard}
\end{figure*}

\subsection{Improved Token Merging through Similarity Distribution Analysis}
\label{subsec:adaptive}

We begin by reviewing the ToMe algorithm. The image is first divided into strides of size $s_x\times s_y$, where one token in each stride is chosen as the destination token. Stride is a partition that helps organize the image into small sections, allowing selective merging of tokens within each section. Depending on the configuration, the destination token can either be selected randomly or always as the top-left token within the stride, while the remaining tokens are considered source tokens. The similarity between each source and destination token pair is then calculated, and each source token is assigned to its most similar destination. 

During inference, in each transformer layer within the $D_1$ and $U_1$ blocks, a fixed number of the most similar tokens, $r$, are merged at each timestep. However, this approach does not consider the distribution of similarities between token pairs, leading to the merging of a predefined number of tokens regardless of how similar they are. As a result, after merging the most similar tokens, less similar pairs may also be merged, potentially leading to poor representative tokens. On the other hand, if more tokens are highly similar, the predetermined value of $r$ limits the number of tokens that can be merged, missing an opportunity to further increase speed without sacrificing quality.

Figure~\ref{fig:adaptive} illustrates two distinct scenarios that demonstrate the limitations of a fixed merging rate. In this example, the image is divided into strides of size $2 \times 2$, where the top-left token in each stride is designated as the destination token, and the remaining tokens within the stride are considered source tokens. In (A), most tokens have low similarity, which means that using a constant merging rate can lead to merging dissimilar tokens, resulting in information loss. Only highly similar tokens are merged by introducing a similarity threshold, thus preserving quality. In (B), it shows a contrasting scenario, typical of early denoising steps, where most tokens have high similarity. A constant merging rate fails to capitalize on the opportunity to merge more tokens without affecting quality. Setting a similarity threshold in this scenario allows for merging a larger number of tokens, significantly speeding up the inference process. 

It is important to note that the scenarios depicted in Figure~\ref{fig:adaptive} are provided to illustrate the challenges associated with fixed merging rates intuitively. While these examples are designed for conceptual clarity, the tensor representations in the latent space lack specific, interpretable meaning for human observation. Nonetheless, the underlying similarity relationships between tokens in the latent space remain valid, and the challenges demonstrated in the figure accurately reflect the dynamics observed during the denoising process. This illustrative approach is employed to enhance understanding of the problem while acknowledging the abstract nature of the latent space.

To address these challenges, we propose an adaptive strategy that dynamically determines the number of tokens to merge based on their similarity distribution. Instead of using a fixed merging number $r$, we introduce a similarity threshold $t$, merging all token pairs with a cosine similarity greater than $t$. This approach ensures that fewer tokens are merged when similarities are low, reducing information loss, while more tokens are merged when similarities are high, enhancing computational efficiency without compromising quality. Importantly, the threshold $t$ introduces a trade-off: higher values improve quality at the expense of speed, while lower values favor speed over quality.


\subsection{Reducing Redundant Computations through Token Pair Caching}
\label{subsec:caching}

Each step of the Token Merging algorithm within the transformer blocks of the U-Net architecture involves significant computational processing, including constructing a bipartite graph between source and destination tokens, calculating a similarity matrix using cosine similarity to measure token similarity, and determining indices for merging between paired source and destination tokens. The most similar source and destination tokens are then connected, and the top $r$ pairs are selected and merged by averaging their features. These operations are computationally expensive and, as our analysis shows, often redundant. It is well-established that changes during different U-Net inference steps are generally smooth[\citenum{wimbauer2024cache}]. Furthermore, our experiments indicate that tokens passed to consecutive transformer blocks exhibit only minor variations, signifying that individual tokens undergo small changes across inference iterations. Specifically, we compute the Jaccard distance between token pairs from neighboring inference steps. For each step \( n \), we create two sets, \(A_n\) and \(A_{n+1}\), which contain all source-destination pairs at steps \(n\) and \(n+1\), respectively. The Jaccard distance between these sets is then calculated using the formula:

\[
\text{JaccardDistance}(A_n, A_{n+1}) = 1 - \frac{|A_n \cap A_{n+1}|}{|A_n \cup A_{n+1}|}
\]

Experimental results from tests conducted on 1000 images reveal that the differences between these token pairs are statistically insignificant, as shown in Figure~\ref{fig:jaccard}. This insight allows us to reuse the token pairs identified for merging across consecutive iterations. We propose using checkpoints, where we compute the indices for merging at each checkpoint. This limits the computation of merging steps—from identifying source and destination tokens to their final merging—only to specific inference steps within each transformer layer. This approach substantially reduces redundant computations without compromising output quality.

Figure ~\ref{fig:graphical_abs} illustrates the caching mechanism as implemented in $D_1$, but our method applies to all the transformer blocks within both $D_1$ and $U_1$. These blocks are critical components of the architecture, responsible for processing high-dimensional latent representations of the image. Given the dense token representations in these layers, they are particularly computationally expensive. In the figure, solid lines represent token pairs computed at specific checkpoints, indicated by blue boxes. The similarity matrix is calculated at these checkpoints, and new token pairs are identified for merging. Conversely, dotted lines denote cached token pairs derived from the most recently computed checkpoint, as shown in gray boxes. This approach avoids repeated computations by leveraging the inherent smoothness of token changes across timesteps, where changes between adjacent steps are minimal.

\section{Implementation Details}
We observed better results in image generation with a stride size of $2\times2$, so we fixed the stride size at $2\times2$. Additionally, we applied our method only to the topmost encoder and decoder layers of the U-Net architecture (i.e., \( D_1 \) and \( U_1 \) in Figure \ref{fig:graphical_abs}). In these layers, the computational cost is higher, making it a candidate for applying our merging method. As mentioned in [\citenum{li2024snapfusion}], the majority of the computational load in a transformer lies in the self-attention mechanism, and our proposed model performs merging in this layer. Finally, we removed the randomness in selecting the destination token within a $2\times2$ window. We always chose the top-left token as the destination token.

\section{Experiments}

\begin{table}[h!]
\centering
\caption{Performance of Our Method Across Different Threshold Values ($t$) on the ImageNet Dataset}
\label{tab:param_setting}
\begin{tabular}{|c|c|c|c|c|}
\hline
\textbf{Threshold} & \textbf{FID} & \textbf{Average Time(s)} & \textbf{PSNR} & \textbf{SSIM} \\ \hline
0.4 & 35.28 & 6.07 $\pm$ 0.007 & 27.90  & 0.191 \\ \hline
0.5 & 35.46 & 6.07  $\pm$ 0.004& 27.909 & 0.208 \\ \hline
0.6 & 35.56 & 6.10 $\pm$ 0.005 & 27.908 & 0.218 \\ \hline
0.7 & 34.30 & 6.23 $\pm$ 0.002 & 27.910 & 0.234 \\ \hline
0.8 & 33.80 & 6.58 $\pm$ 0.004 & 27.904 & 0.239 \\ \hline
0.9 & 33.42 & 6.92 $\pm$ 0.003 & 27.907 & 0.238 \\ \hline
1.0 & 33.66 & 7.61 $\pm$ 0.001 & 27.905 & 0.241 \\ \hline
\end{tabular}
\end{table}

To evaluate the performance, we utilize Stable Diffusion v1.5 to generate 2000 images at a resolution of $512\times512$ pixels from the ImageNet-1k dataset with a Tesla V100S GPU. This involves generating two images per class for all 1000 classes, using 50 PLMS diffusion steps, with the CFG scale set to 7.5 and the guidance rescale set to 1. To assess image quality, we calculate the FID scores between our 2000 generated images and 5,000 validation images from the ImageNet-1k dataset. Additionally, we use the same prompt as the original work, which is specified in their \#55 GitHub issue, though not directly mentioned in the paper. The prompt for each sample is: \textit{"A high-quality photograph of a $class name$}.” We measure the average time to generate all 2000 samples for speed evaluation.

Our main experiments are divided into two sections, corresponding to the two methods we propose. First, we identify the optimal threshold $t$ for adaptive merging as described in the methodology. Within the threshold range of $0.4$ to $0.6$, the changes are minimal because most source tokens have a cosine similarity of $0.6$ or higher with their most similar destination tokens. The first notable change appears at threshold $0.7$, which we select for our merging experiments across different checkpoints. As hypothesized in the methodology, we observe that with increasing threshold values, the FID decreases as fewer tokens merge, resulting in reduced information loss but increased average processing time. Changes in PSNR are negligible due to the robustness of pixel-level intensity reconstruction, which is not highly sensitive to token merging in our adaptive framework. However, the SSIM metric exhibits a rising trend with increasing thresholds, indicating improved structural alignment between the generated and original images produced by stable diffusion v1.5. This improvement suggests that higher thresholds preserve the spatial and structural consistency of features, as fewer token mergers reduce the likelihood of spatial distortions or blending errors. Consequently, this ensures that the generated images more accurately retain the intricate details and textures of the original dataset, as measured by SSIM.

After selecting the optimal threshold, we evaluate six configurations, detailed in Table \ref{tab:checkpoints_values}. Each configuration has a different checkpoint distribution across inference steps. For example, CONFIG\_1 has more checkpoints in the early inference steps, while CONFIG\_3 has more in later steps. Among these, CONFIG\_3 demonstrates superior performance in terms of both time and FID. This result aligns with our expectations, as indicated by the Jaccard distance in Figure \ref{fig:jaccard}, which is higher around the $8th$, $11th$, and $13th$ steps and also shows increased differences in indices pair in the final inference steps. Consequently, CONFIG\_3 performs better overall with more checkpoints in these points.

Our final results, presented in Table \ref{tab:final_results}, highlight the effectiveness of our method. Our approach outperforms the ToMe method in both time and FID.

We also reported PSNR and SSIM as additional quality measures. PSNR quantifies the quality ratio between the original and generated images, while SSIM evaluates their structural similarity. To compute these metrics, we paired the generated images with their corresponding original images for each class and calculated the average value for each pair.   




\begin{table}[h]
\centering
\caption{Checkpoints Name Convention}
\label{tab:checkpoints_values}
\begin{tabular}{|c|l|}
\hline
\textbf{Name} & \textbf{Checkpoints} \\
\hline
CONFIG\_1 & [0,1,2,3,5,10,15,25,35] \\
CONFIG\_2 & [0,10,11,12,15,20,25,30,35,45] \\
CONFIG\_3 & [0,8,11,13,20,25,30,35,45,46,47,48,49] \\
CONFIG\_4 & [0,9,13,14,15,28,29,32,36,45] \\
CONF\_ADAPTIVE & [0,1,5,7,10,12,15,35,40,45,46-51] \\
CONFIG\_Five & [0,5,10,15,20,25,30,35,40,45,50] \\
\hline
\end{tabular}
\end{table}

\begin{table}[h!]
\centering
\caption{Comparison of different methods, thresholds, cache settings, and their respective results in terms of time and FID scores.}
\label{tab:diff_thresholds}
\begin{tabular}{|c|c|c|c|c|}
\hline
\textbf{Method/Model} & \textbf{t / r} & \textbf{Cache Setting} & \textbf{Time} & \textbf{FID} \\ \hline
Ours & 0.7 & CONF\_1 \ref{tab:checkpoints_values} & 6.18 $\pm$ 0.02 & 36.14 \\ \hline
Ours & 0.7 & CONF\_2 & 6.13 $\pm$ 0.001 & 34.33 \\ \hline
\rowcolor{lightgray} Ours & 0.7 & CONF\_3 & 6.09 $\pm$ 0.001 & 34.05 \\ \hline
Ours & 0.7 & CONF\_4 & 6.12 $\pm$ 0.001 & 34.82 \\ \hline
Ours & 0.7 & CONF\_ADAPTIVE & 6.19 $\pm$ 0.001 & 35.56 \\ \hline
Ours & 0.7 & CONF\_FIVE & 6.14 $\pm$ 0.001 & 35.20 \\ \hline
Difusion\_model & - & BASELINE & 7.61 $\pm$ 0.001 & 33.66 \\ \hline
ToMe & 0.5 & No cache & 6.39 $\pm$ 0.006 & 34.16 \\ \hline
\end{tabular}
\end{table}

\begin{table}[h!]
\centering
\caption{Quantitative results on generating 2000 images of ImageNet dataset}
\label{tab:final_results}
\begin{tabular}{|c|c|c|}
\hline
\textbf{Model} & \textbf{FID} & \textbf{Average Time(s)} \\ \hline
Baseline[1] & 33.66 & 7.61$\pm$0.001 \\ \hline
ToMe[1]     & 34.16 & 6.39 $\pm$ 0.006 \\ \hline
Ours        & 34.05 & 6.09$\pm$0.001 \\ \hline
\end{tabular}
\end{table}


\section{Conclusion}
In this paper, we propose a novel method, called cached adaptive token merging(CA-ToMe), which exploits the inherent redundancy within images and the temporal redundancy present between consecutive steps to accelerate the inference process while preserving the quality of generated images. Our approach introduces an adaptive merging mechanism, wherein a threshold is established to determine the similarity tolerance for tokens selected for merging. Building upon the observation that token pairs selected for merging exhibit similar patterns across steps, and that the output of the denoising U-Net blocks displays smoothness, CA-ToMe endeavors to cache and reuse merging token pairs in subsequent steps. By doing so, our method achieves a superior acceleration rate compared to traditional token merging approaches, while maintaining the image quality.

\bibliographystyle{IEEEtran}

\bibliography{references}

\end{document}